%% file: acl_latex.tex
\documentclass[11pt]{article}

\usepackage[preprint]{acl}
\usepackage{dblfloatfix}
\usepackage{times}
\usepackage{latexsym}
\usepackage{amsmath} 
\usepackage{amssymb}
\usepackage[T1]{fontenc}

\usepackage[utf8]{inputenc}
\usepackage{listings}
\usepackage{xcolor}

\usepackage{microtype}
\usepackage{booktabs}
\usepackage{inconsolata}

\usepackage{graphicx}

\usepackage[table]{xcolor}
\definecolor{GrayRow}{HTML}{F0F0F0}
\newcommand{\highlightrow}{\rowcolor{GrayRow}}
\definecolor{BlueRow}{HTML}{F0F8FF}
\newcommand{\resultrow}{\rowcolor{GrayRow}}

\usepackage{enumitem}
\usepackage{pgfplots}
\usepackage{tabularx}
\usepackage{caption}
\usepackage{subcaption}
\usepackage{bbm}
\usepackage{url}
\usepackage{amsmath} 
\usepackage{pifont}
\usepackage{arydshln}
\usepackage{makecell}
\usepackage{booktabs}
\usepackage{multirow} 
\usepackage{algorithm}
\usepackage{amssymb}
\usepackage{xcolor}
\usepackage{tcolorbox}
\usepackage{bbding}
\usepackage{siunitx}
\usepackage{newtxmath}
\usepackage{nicefrac}
\usepackage{pgfplots}
\usepackage{algpseudocode}
\usepackage{bm}

\usepgfplotslibrary{groupplots}
\pgfplotsset{compat=1.18}
\title{Don't Wait to Reply: Towards Responsive yet Thoughtful Dialogue\\through Proactive Thinking}

\author{
 \textbf{Ante Wang}\textnormal{\textsuperscript{$\diamondsuit$}}\thanks{These authors contributed equally.},
 \textbf{Jiaqi Fu}\textnormal{\textsuperscript{$\clubsuit$}}\footnotemark[1],
 \textbf{Xuanyi Chen}\textnormal{\textsuperscript{$\clubsuit$}},
 \textbf{Ruotian Ma}\textnormal{\textsuperscript{$\spadesuit$}},
 \textbf{Zhaopeng Tu}\textnormal{\textsuperscript{$\spadesuit$}},\\
 \textbf{Weizhi Ma}\textnormal{\textsuperscript{$\diamondsuit$}} \textnormal{and}
 \textbf{Yang Liu}\textnormal{\textsuperscript{$\clubsuit$,$\diamondsuit$}}
\\
 \textsuperscript{$\clubsuit$}Dept. of Comp. Sci. \& Tech., Institute for AI, Tsinghua University\\
 \textsuperscript{$\diamondsuit$}Institute for AI Industry Research (AIR), Tsinghua University \quad \textsuperscript{$\spadesuit$}Tencent \quad
}

\begin{document}
\maketitle
\begin{abstract}

Thinking has emerged as a critical capability for Large Language Models (LLMs) tackling complex tasks. However, its reactive nature, where reasoning is passively triggered only upon receiving a user response, inevitably introduces latency that compromises conversational fluidity. This stands in sharp contrast to human dialogue, where speakers proactively anticipate and plan future content during natural pauses to ensure seamless interaction. To bridge this gap, we propose Proactive Thinking, a framework that empowers models to pre-compute potential response elements during conversational downtime instead of waiting idly for the next input.
We then introduce a training-free baseline that can think ahead by anticipating future states, balancing efficiency and quality through speculative continual thinking. To evaluate this approach in practice, we adapt three benchmarks of varying complexity into time-aware environments that simulate real-time conversational flow. We demonstrate that proactive thinking effectively improves interaction efficiency without compromising performance. Ultimately, this work advocates for a fundamental shift toward more intelligent, anticipatory, and real-time conversational AI.

\end{abstract}

\section{Introduction}
The appeal of natural conversation lies in its interactive flow. A crucial component of this flow is the inter-speaker interval, commonly around 200 ms for reactions~\cite{heldner2010pauses}. Although this duration varies considerably across cultures and topics, it seldom lasts longer than four seconds. Longer pauses usually result in an awkward silence~\cite{mclaughlin1982awkward}, making participants uncomfortable.

The rise of Large Language Models (LLMs, \citealt{achiam2023gpt,touvron2023llama,yang2025qwen3,liu2025deepseek}) has reshaped interactions that previously occurred mainly between humans. People now converse with LLM agents to accomplish various tasks, a capability enabled by the models' sophisticated linguistic skills, learned from vast corpora that include human dialogue. Recently, conversational abilities have been advanced further through chain-of-thought reasoning~\cite{wei2022chain,yao2022react,bhaskar2025language}. This ``think-before-respond'' paradigm significantly enhances response quality by improving the understanding of user intent~\cite{feng2025improving}, enabling more efficient goal achievement~\cite{lai2025doctor}, and incorporating safety considerations~\cite{jiang2025safechain}.

However, this paradigm inherently introduces latency due to its reactive nature. LLMs always begin thinking only after their dialogue turn starts, often consuming hundreds or even thousands of tokens for reasoning before generating a final response. This delay severely limits the applicability of LLM agents in live conversations and other real-time interactive scenarios, where both response quality and low latency are crucial for user experience.

A natural direction for addressing latency is to accelerate the reasoning process itself. Prior work has explored several avenues. For instance, techniques such as model distillation and specialization create smaller, faster models~\cite{hsieh2023distillingstepbystepoutperforminglarger, zhao-etal-2024-probe, li2024mixeddistillationhelpssmaller}. Other approaches include adaptive thinking or adding length penalties to encourage shorter reasoning chains~\cite{zhang2025adaptthink,kang2025c3ot,aggarwal2025l1}. Alternatively, token-level caching and speculative decoding partially address the problem at a system level~\cite{zhou2024distillspecimprovingspeculativedecoding, huang2025selectkdselectivetokenweightedknowledge}. Though effective, these approaches either sacrifice reasoning performance or require substantial and costly engineering effort.

\begin{figure*}[t]
\centering
\includegraphics[width=1\linewidth]{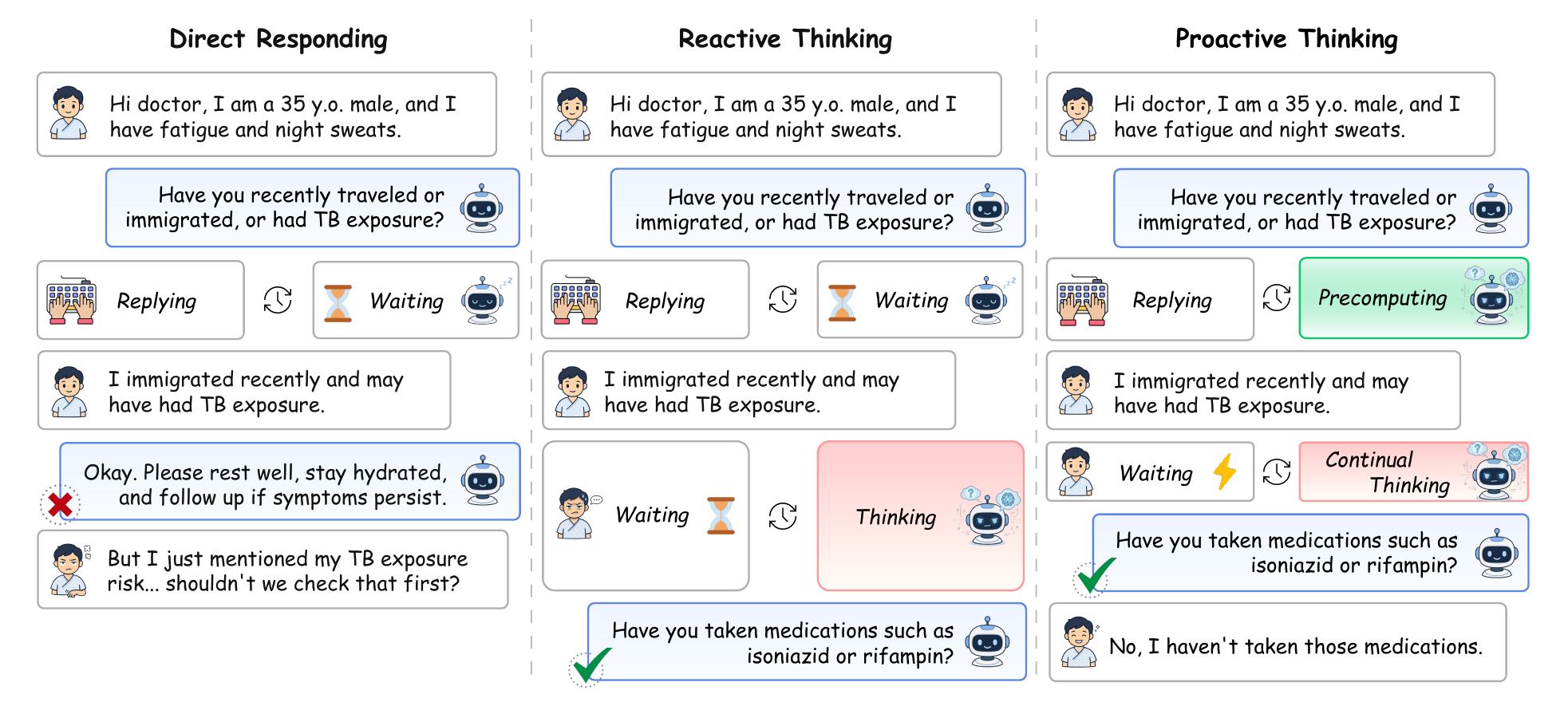}
\vspace{-2em}
\caption{
Comparison of direct responding, reactive thinking and proactive thinking paradigms.
}
\label{fig:comparison}
\end{figure*}

To address this limitation, we propose a paradigm shift by introducing \textbf{Proactive Thinking}. This approach simulates human social cognition by utilizing the natural idle periods within a dialogue to perform reasoning for future states in advance~\cite{levinson2016turn}. When a user reply arrives, the LLM can respond immediately with precomputed results. This effectively transforms reasoning from a passive computation into a preparatory resource.
However, determining what an LLM should reason about during these windows and how to effectively utilize that information while maintaining performance remains an open question.

To answer this, we introduce a training-free framework consisting of two mechanisms: \textit{precomputation via anticipated rollouts} and \textit{speculative continual thinking}. By anticipating potential user responses and reasoning through them, the LLM can directly reuse precomputed results if the actual input matches a speculation. In cases where the input diverges, the LLM performs continual thinking to prevent performance degradation. Since LLMs often struggle to decide optimally when to continue thinking, we further integrate speculative decoding~\cite{leviathan2023fast} into our framework. This system determines the appropriate amount of precomputed content to reuse through self-verification with theoretical performance guarantees.

To evaluate this paradigm, we introduce Time-Aware Interaction Environments adapted from 20 Questions~\cite{qian2025userrltraininginteractiveusercentric}, AgentClinic~\cite{schmidgall2025agentclinicmultimodalagentbenchmark}, and IN3~\cite{qian2024tell}. Unlike existing benchmarks, our environment unifies token generation and human typing speeds to estimate the dialogue countdown and response latency, allowing for the simulation of realistic time consumption. These datasets feature varying levels of complexity regarding potential user responses to test the robustness of our proposed methods.

Experimental results across various LLMs demonstrate that proactive thinking significantly reduces latency compared to the reactive thinking baseline, without compromising performance. Further analyses indicate that our method is robust across different environmental setups and hyperparameters. However, the ability to accurately anticipate user replies and utilize them more efficiently remains a key area for future improvement.

In summary, this paper makes the following contributions:
\begin{itemize}[leftmargin=*]
\item We propose proactive thinking, a paradigm shift from conventional reactive thinking that advances the interaction capability of LLMs by reducing response latency via precomputation during dialogue idle time.
\item We propose a training-free approach to implement proactive thinking through precomputation via anticipated rollouts and speculative continual thinking, serving as a strong baseline for future research.
\item We develop time-aware interaction environments of varied difficulty by adapting existing datasets to support the standardized evaluation of proactive thinking.
\item We validate the effectiveness of proactive thinking and analyze its critical factors, providing insights for further optimization of this paradigm.
\end{itemize}

\section{Preliminaries}

In this research, we focus on reasoning-intensive multi-turn dialogue tasks. Unlike open-domain chitchat, these tasks require the LLM to achieve a specific goal through strategic interaction. For example, in a clinical consultation, the model must systematically inquire about symptoms to efficiently reach an accurate diagnosis.
We first formalize the conventional reactive thinking paradigm (\S \ref{sec:react}) and then present a pilot study highlighting its inherent tradeoff between response quality and interaction latency (\S \ref{sec:pilot}).

\subsection{Reactive Thinking}

\label{sec:react}
Multi-turn dialogue is modeled as a sequence of interactions between an LLM $\pi_{\theta}$ and a user. At each turn $t$, the model receives a user input $o_t$ and generates a response $a_t$. Let \(\tau_{<t} = (o_1, a_1, \dots, o_{t-1}, a_{t-1})\) represent the dialogue exchange up to the current turn.

To improve the quality of $a_t$, a common practice is to predict a reasoning trace $r_t$ before producing the response $a_t$:
\[
r_t, a_t = \pi_{\theta}(\tau_{<t}, o_t).
\]

As the reasoning $r_t$ is conditioned on the most recent user input $o_t$, the model remains idle while the user is composing their message, and only begins the computation for $r_t$ after $o_t$ is received.

\subsection{The Performance and Latency Tradeoff}
\label{sec:pilot}

\input{assets/pilot_study}

We conducted a pilot study on AgentClinic~\cite{schmidgall2025agentclinicmultimodalagentbenchmark} to analyze how reasoning budgets impact dialogue performance. We evaluated Qwen3~\cite{yang2025qwen3} models by varying the maximum allowed number of reasoning tokens during each turn of the conversation.

As illustrated in Figure \ref{fig:pilot_study}, task accuracy scales positively with the depth of reasoning. Both models achieve their best performance at an average of around 400 tokens. However, this performance gain comes at the cost of significantly increased response time. Achieving peak accuracy requires about 4 seconds of reasoning after the user sends a message. This leads to high latency that can make the interaction feel sluggish and less responsive.

This bottleneck arises because the model follows a rigid sequence: it waits for the user input before beginning its reasoning process. This observation leads to our central research question: \textbf{can reasoning be decoupled from the immediate user input to reduce latency?}


\section{Proactive Thinking}

In this section, we first outline the conceptual framework of proactive thinking (\S \ref{sec:framework}). Next, we introduce a training-free baseline designed to theoretically reduce latency without compromising performance (\S \ref{sec:baseline}).

\subsection{Conceptual Framework}
\label{sec:framework}

In contrast to reactive thinking (see \S\ref{sec:react}), our method proactively thinks ahead during the interval between the LLM’s response \(a_{t-1}\) and the user’s incoming reply \(o_{t}\). During this period, the LLM precomputes a reasoning trace \(z_t\) based on the historical context:
\[
z_t \sim \pi_{\theta}(\tau_{<t}).
\]

Upon receiving the true reply \(o_{t}\), the precomputation is immediately interrupted if it has not yet finished. The LLM then generates its next response \(a_{t}\) conditioned on the precomputed result $z_t$:
\[
e_{t}, a_{t} = \pi_{\theta}(\tau_{<t}, o_{t}, z_t),
\]
where \(e_{t}\) is an optional continuation of the thinking process. 
To account for constraints such as limited waiting time, model capability, or task complexity, this design ensures robustness even when the precomputed \(z_t\) does not perfectly align with the actual reply \(o_t\). While it shares similarities with the reactive reasoning process, it effectively utilizes \(z_t\) as a potentially useful preparatory context. If the necessary reasoning content is already covered within \(z_t\), the remaining reasoning workload is minimized, thereby reducing generation latency.

\subsection{A Training-Free Baseline}
\label{sec:baseline}

Implementing proactive thinking introduces two critical challenges: determining what the LLM should precompute, and adaptively reusing those precomputed results during continual thinking. We address these challenges via \textit{anticipated rollout precomputation} and \textit{speculative continual thinking}, respectively. We detail this approach in Algorithm~\ref{algo}.

\paragraph{Precomputation via anticipated rollouts}
\label{sec:speculative_rollouts}

The main challenge of proactive thinking is the absence of the true user reply $o_t$ during the precomputation window. Nevertheless, LLMs trained on massive dialogue corpora inherently develop the capacity to anticipate probable user utterances given a conversational context. This predictability creates an opportunity for computational acceleration: an LLM can precompute the thinking process for the most probable user replies while waiting for actual feedback.

Formally, at the $t$-th dialogue turn, immediately after generating response $a_{t-1}$ but before observing the next user utterance $o_t$, we prompt the model to hypothesize $k$ plausible user replies, denoted as $\hat{O}_t = \{\hat{o}_t^{(i)}\}_{i=1}^k$. For each hypothesized reply, the model precomputes a corresponding anticipated rollout containing the follow-up reasoning and response $z_{t}^{(i)}$:
\begin{align*}
\hat{O}_t &= \pi_{\theta}(P, \tau_{<t}),\\
z_{t}^{(i)} &= \pi_{\theta}(\tau_{<t}, \hat{o}_t^{(i)}),
\end{align*}
where $P$ denotes the prompt utilized for hypothesis generation, respectively. The resulting set of hypothesis-rollout pairs $\{(\hat{o}_t^{(i)}, z_t^{(i)})\}_{i=1}^{k}$ constitutes the precomputed cache $\mathcal{Z}_t$. Note that precomputation is bounded by the real-world idle interval; thus, $\mathcal{Z}_t$ may be incomplete, containing only partially generated content.

\paragraph{Speculative continual thinking}
\label{sec:adaptive_continuation}

Upon receiving the actual user reply $o_{t}$, a naive approach would be to directly append $\mathcal{Z}_t$ to the context. However, without explicit fine-tuning, LLMs struggle to natively balance the reuse of mismatched precomputed elements with fluid continued thinking, often leading to performance degradation. To resolve this, we treat $\mathcal{Z}_t$ as a draft and employ a modified speculative decoding mechanism, allowing the model to reuse valid reasoning content via self-verification.

We first select the most relevant hypothesis using a lightweight similarity metric:
\[
k^* = \operatorname{arg\,max}_{k} \; s\left(o_{t}, \hat{o}_t^{(k)}\right),
\]
where $s(\cdot,\cdot)$ denotes a similarity score (e.g., string matching or embedding similarity). For the selected precomputed rollout $z^{k^*}_t$, each drafted token $z^{k^*}_{t,i}$ is accepted with an adjusted probability:
\[
\alpha_{i} = \min\Biggl(1,\; \ell \cdot \frac{\pi_{\theta}(z^{k^*}_{t,i} \mid \tau_{<t}, o_{t}, z^{k^*}_{t,<i})}{\pi_{\theta}(z^{k^*}_{t,i} \mid \tau_{<t}, \hat{o}_t^{(k^*)}, z^{k^*}_{t,<i})}\Biggr),
\]
where the lenience parameter $\ell \ge 1$ relaxes the strict speculative acceptance condition to balance generation quality and reuse efficiency~\cite{liu2025spec}. The verified content prefix $z^{k^*}_{t,<j}$ is determined by the first token position $j$ that is rejected by this criterion.

The final response is then generated by continuing from this accepted prefix rather than restarting reasoning from scratch:
\[
e_t, a_t \sim \pi_{\theta}(\tau_{<t}, o_t, z^{k^*}_{t,<j}).
\]

The continuation trace $e_t$ is typically shorter than the standard reasoning trace required by reactive thinking, thereby yielding latency reductions. Moreover, the verification step guarantees that all reused content remains valid conditioned on the actual observation $o_t$, preserving task performance.

\paragraph{Discussion}
The efficacy of this training-free strategy is fundamentally contingent upon the model's capacity to anticipate the distribution of possible user utterances. If the model fails to capture the actual user response, the utility of the precomputed rollout set $\mathcal{Z}_t$ diminishes because less content can be reused. This highlights a critical dependency on the entropy of the conversational context. In structured or task-oriented scenarios where conversations follow predictable, logical trajectories, the model can effectively span the most probable outcomes with a small $k$, resulting in substantial latency reductions. Conversely, within open-ended or high-entropy exchanges, the vast state space of potential user replies makes it difficult to generate a matching hypothesis, which may render precomputation less effective. As a result, while this approach offers robust acceleration, the performance gains are most pronounced in focused interactions where the range of plausible user feedback is constrained.

\section{Time-Aware Interaction Environment}

Standard multi-turn benchmarks predominantly evaluate the semantic quality of agent responses while overlooking the temporal dynamics inherent in real-world dialogue. We address this gap by proposing a framework that simulates time-aware interactions, enabling the evaluation of an agent's ability to utilize idle time for proactive thinking. 

\subsection{Benchmark Compositions}
We utilize three interactive tasks with varying levels of complexity in anticipating user replies:
\begin{itemize}[leftmargin=*]
    \item \textbf{20 Questions:} This is a cooperative game where a player attempts to guess a secret object by asking up to 20 yes-or-no questions. It is frequently used to evaluate a model's deductive reasoning, strategic inquiry, and information-seeking capabilities. We use 401 cases from \citet{qian2025userrltraininginteractiveusercentric} for evaluation.
    \item \textbf{AgentClinic}~\cite{schmidgall2025agentclinicmultimodalagentbenchmark}: This environment simulates a dynamic medical setting by framing diagnostic tasks as multi-turn dialogues, evaluating LLMs on sequential clinical decision-making rather than static question-answering. We adopt the set based on MedQA~\cite{jin2020diseasedoespatienthave}, which comprises 214 patients.
    \item \textbf{IN3}~\cite{qian2024tell}: Intent understanding is a fundamental capability in dialogue systems. This benchmark evaluates an agent's capacity to infer implicit user intentions through explicit multi-turn interactions. We adopt a subset of 420 cases from \citet{qian2025userrltraininginteractiveusercentric}, covering diverse domains such as cookery, arts, and programming.
\end{itemize}
Following prior work, we employ an LLM as a user simulator to ensure controllable and reproducible interactions. The specific prompt configurations for these simulators are detailed in \S \ref{app:prompt}.

\subsection{Temporal Modeling}
\label{sec:temporal_modeling}
Because of the significant discrepancy between LLM inference throughput and human cognitive processing speeds, measuring simulated wall-clock time is unsuitable. Furthermore, hardware heterogeneity and varying software environments often compromise reproducibility. Therefore, we model time based on standardized token generation rates and human typing speeds. We introduce two relevant metrics for this simulation:

\begin{itemize}[leftmargin=*]
    \item \textbf{Response Latency:} This represents the delay between the completion of user input and the delivery of the LLM's response. As illustrated in Figure \ref{fig:pilot_study}, generation time grows almost linearly with the number of generated tokens across different LLMs on AgentClinic, a pattern we also observe in the other datasets. We define this latency as $\frac{N_{out}}{v}$, where $N_{out}$ is the number of generated tokens and $v$ is a constant token generation rate per second. 
    \item \textbf{Dialogue Countdown:} This represents the interval during which the user processes information and composes a reply, providing a potential thinking window for the agent. This window comprises a fixed cognitive delay $d$ and a variable composition time based on the user's typing speed $v_{u}$. The total window is modeled as $d + \frac{N_{in}}{v_{u}}$, where $N_{in}$ represents the number of tokens in the user's reply. We calibrate $v_{u}$ to align with standard human typing rates of 30 to 60 words per minute~\cite{dhakal2018observations,palin2019people}. 
\end{itemize}

While this framework is currently implemented for text-based communication, the underlying logic is natively extensible to other interaction modalities, such as speech.

\input{assets/main_exp_cleaned}

\section{Experiment}
\label{sec:experiment}

\subsection{Setup}

\paragraph{Models and Hyperparameters}
By default, we set $v=100$, $\tau=0.5$, and $v_u=1$ across all environments, based on preliminary experiments and empirical findings. We also explore the influence of these parameters across different simulation scenarios in our analysis. For user simulation, we employ gpt-oss-120B across all environments. To evaluate proactive thinking, we compare it against conventional direct responding and reactive thinking across various LLMs, including Qwen~\cite{yang2025qwen3}, Gemma~\cite{team2024gemma}, and Llama~\cite{touvron2023llama}. The specific prompts used in this study are provided in Appendix~\ref{app:prompt}. All models are deployed using vLLM on four A100 GPUs with a generation temperature of $0.5$ to encourage response diversity. For proactive thinking, we utilize the small embedding model trained by \citet{gao2021simcse} for rollout selection and set $\gamma=5$ for self-verification. Due to the limited sizes of AgentClinic, we report the average of four runs to ensure the reliability of our results.

\paragraph{Evaluation Metrics}
We evaluate each method based on task performance and dialogue latency. Following prior work, task performance is measured by Accuracy and the number of turns (\#Turns). Accuracy indicates whether the model successfully achieves the objective (e.g., making an accurate diagnosis). The computational details of this metric for each environment are described in \S\ref{app:eval}. The \#Turns metric measures interaction efficiency, where a lower number indicates higher efficiency. For dialogue latency, we report the average latency (in seconds) per dialogue turn, computed within the simulated environment defined in \S\ref{sec:temporal_modeling}.

\subsection{Main Results}
\label{sec:main_results}

Table~\ref{tab:main-results} shows the test results of different methods in our time-aware interaction environments.
Compared with direct response, reactive thinking consistently improves task performance, yielding higher accuracy or fewer interaction turns across all models and datasets. However, it incurs a significant latency penalty, echoing the findings of our pilot study.

With proactive thinking, we observe a \textit{substantial reduction in latency} compared to reactive thinking, while \textit{maintaining competitive task performance}. This improvement occurs because proactive thinking allows the model to reuse reasoning traces from precomputation, thereby reducing the number of tokens that need to be generated. Additionally, because the reused content has already undergone verification, it avoids the noise that typically leads to performance degradation.

We observe that the gains in latency reduction diminish when transitioning from 20 Questions to AgentClinic, and further to IN3. This trend stems from the inherent complexity of anticipating user responses, as detailed in our expanded analysis in \S \ref{app:app_exp}. Intuitively, a patient's response is more diverse than that of the oracle in 20 Questions, while IN3 introduces an even greater challenge. Nevertheless, our method consistently preserves task performance, validating its robustness.

\subsection{Analyses}
\label{sec:analysis}

We conduct further analyses on several key factors to better understand the underlying mechanism of proactive thinking. By default, our experiments are evaluated on AgentClinic using Qwen3.6-27B. Due to space constraints, additional analyses and detailed case studies are provided in \S \ref{app:app_exp}.

\input{assets/env_analysis}

\paragraph{Performance in different environment setups}

User typing speed and model generation speed can vary significantly across different environments. To quantify their influence on the final latency, we adjust their corresponding hyperparameters; the results are presented in Figure~\ref{fig:env_analysis}. As expected, latency increases alongside user typing speed. This occurs because a faster user reply shortens the dialogue countdown, which can prevent the anticipated rollout from completing its generation. A faster token generation speed reduces latency for both the proactive and reactive thinking approaches. Nevertheless, our method consistently maintains an obviously lower latency, demonstrating its generalization capabilities across diverse environments.

\input{assets/reply_num}

\paragraph{Effectiveness of anticipating the user responses}
In Table~\ref{tab:k_analysis}, we evaluate the impact of hypothesis generation on performance. We first analyze the 20 Questions dataset, which allows for precise adjustment of the gold user reply's recall. We compare our method against two variants: one setting $k=1$ and another utilizing an incorrect user reply for precomputation. The results demonstrate that successfully anticipating user responses fosters a more consistent reasoning process, which enhances the token acceptance rate and reduces latency.
We observe a similar trend on AgentClinic. However, since LLMs can be highly sensitive to verbosity, phrasing variations can significantly alter the output distribution even with semantically identical content~\cite{zhou2024distillspec}. Thus, there remains considerable room for improvement through further optimization. 
We further explore alternative baselines, such as employing an empty string or directly copying the previous model response as the anticipated rollout. Both configurations perform substantially worse than our approach, validating the necessity of our hypothesis generation step.


\input{assets/gamma}

\paragraph{Robustness to the Lenience $\gamma$}
Typically, $\gamma$ relaxes the acceptance criteria for draft tokens. A larger $\gamma$ allows for a longer reused prefix, thereby reducing latency, though at the potential risk of degrading performance. Surprisingly, as shown in Figure~\ref{fig:gamma}, while we observe a significant reduction in latency with larger values of $\gamma$, there is no corresponding drop in task performance. We attribute this to the fact that the probability discrepancy for tokens that ought to be rejected is often quite large across different contexts. Consequently, $\gamma$ can be selected from a wide range, demonstrating the robustness of our method to this parameter.

\section{Related Work}

\subsection{Think-Before-Response Paradigm}
The concept of Chain-of-Thought (CoT) was pioneered by \cite{wei2022chain}, demonstrating that encouraging models to generate intermediate reasoning steps substantially improves performance on complex tasks. This approach effectively mirrors human ``System 2'' thinking \cite{evans2013dual}, where intricate problems are decomposed into sequential, deliberate steps. Subsequent research has further enhanced this capability by constructing high-quality reasoning chains via rejection sampling \cite{zelikman2022star, yu2024metamath} or tree search methods \cite{guan2025rstar}, significantly boosting performance on reasoning-intensive tasks such as mathematics.

Recent advances have successfully leveraged Reinforcement Learning (RL, \citealt{schulman2017proximal, shao2024deepseekmath}) to scale inference-time computation \cite{guo2025deepseek, team2025kimi, yang2025qwen3}. By incentivizing models to generate longer, more rigorous reasoning traces, these approaches have yielded remarkable performance gains across a broader range of domains, including complex multi-turn dialogues \cite{bhaskarlanguage, qian2025userrltraininginteractiveusercentric, gandhi2026learning}.

However, this paradigm introduces a significant trade-off between task performance and response latency. Although various optimization techniques have been explored to mitigate these delays, such as efficient decoding algorithms \cite{wang2025efficient, shi2025speccot, shen2025codi} or methods to compress thinking length \cite{wu2025unlocking, laaouach2025halt, zhang2025adaptthink, team2025kimi}, they fundamentally operate within a traditional reactive thinking framework. In this work, we propose a paradigm shift toward proactive thinking. This approach is orthogonal to the optimization techniques mentioned above; thus, combining them is expected to unlock further latency reductions and performance gains.

\subsection{Multi-Turn Dialogue Tasks}
Dialogue modeling has long been a foundational challenge in natural language processing (NLP). Early research typically categorized dialogue systems into task-oriented and open-domain tracks \cite{chen2017survey}. While the emergence of LLMs has drastically advanced open-domain chit-chat capabilities \cite{achiam2023gpt}, these models still face challenges in complex, task-oriented scenarios that demand intensive reasoning and multi-step sequential decision-making \cite{li2024mediq, laban2025llms}.

Beyond turn-based text communication, streaming modalities such as spoken dialogue represent another critical frontier \cite{ji2024wavchat}, where user input is received continuously. This continuous flow has catalyzed incremental processing approaches \cite{hastie2012incremental, kennington2025incremental} designed to process information word by word. For instance, \citet{shih2025can} proposed initiating the model's thinking process concurrently while the speaker is still talking. Similarly, \citet{wang2024full} and \citet{zhang2024beyond} allowed systems to continuously receive user inputs and generate responses. However, both frameworks do not investigate the role of internal thinking in such contexts. Alternatively, \citet{xie2025mini} proposed interleaving thinking with response generation, distributing the computational latency across individual tokens to minimize the user's perceived delay.

Despite these advances, these methods are restricted to streaming setups and remain bounded by the reactive thinking paradigm: they wait for (at least partial) user input to trigger reasoning, rather than thinking proactively. As a result, they fail to utilize strategic windows such as turn-taking gaps or initial, low-information filler words. Integrating proactive thinking with incremental processing thus offers a promising synergy to further minimize latency and enhance real-time interaction quality.

\section{Conclusion}
In this work, we proposed proactive thinking to reduce response generation latency in multi-turn dialogues. Unlike standard reactive reasoning, which must wait for the user's latest reply to begin, our approach decouples the thinking process from the sequential interaction trajectory by precomputing thoughts for later use. We introduced a training-free baseline leveraging anticipated rollout for precomputation, alongside a continual-thinking mechanism based on speculative decoding to efficiently reuse precomputed results. To validate our approach, we developed a time-aware interaction environment based on three existing benchmarks of varying difficulty. Our results demonstrate that the proposed method effectively reduces interaction latency without compromising performance.

This study serves as a preliminary exploration of proactive thinking. We believe these results can be further improved through the following directions:
\begin{itemize}[leftmargin=*]
\item \textbf{User intent understanding:} The performance of a proactive agent is bounded by its ability to accurately anticipate future states, which requires a deep understanding of user intents. This is conceptually analogous to predicting state transitions in a Markov decision process~\cite{ha2018world, lecun2022path}.
\item \textbf{Tailored reasoning:} While our training-free baseline is effective, it faces computational inefficiencies due to its handcrafted workflow. This limitation could be addressed through reinforcement learning (RL)~\cite{guo2025deepseek,qian2025userrltraininginteractiveusercentric} to train LLMs to actively explore efficient reasoning strategies optimized for the final goal. Determining how to best guide exploration and design effective rewards remains an important, under-explored question.
\item \textbf{Broader application:} Proactive thinking can be combined with incremental processing techniques to better support streaming dialogue environments. Furthermore, this paradigm can be extended to more diverse interactive environments, such as real-time strategy games~\cite{ma2024large} or autonomous driving~\cite{wang2023drivemlm}, enabling instant, safer, and higher-quality decisions.
\end{itemize}

\section*{Limitations}
This work serves as a preliminary exploration into the novel concept of proactive thinking; however, several limitations must be addressed before the approach can be fully operationalized. First, while proactive thinking successfully reduces latency, it currently incurs higher computational costs due to the necessity of anticipating future states. This overhead remains unavoidable because the system cannot yet perfectly predict every outcome. Future iterations could employ adaptive mechanisms to trigger these predictive processes only at optimal moments, thereby reducing resource consumption.

Second, we have not yet explored training-based approaches, which would likely yield more substantial improvements by optimizing model parameters directly. Pursuing this direction will require significant effort, as existing models lack inherent proactive thinking capabilities. Consequently, a critical next step involves developing effective data generation pipelines to curate large-scale datasets specifically designed to foster this behavior during training.

Finally, the current time-aware interaction environment may not fully capture the complexities of real-world settings, where variables such as human recognition time and typing speed fluctuate. Additionally, our evaluation focuses on idealized time consumption. In practice, deployed performance will depend on engineering refinements, such as the ability to flexibly interrupt precomputations upon receiving a user reply and to allow prefix continuation by reusing the KV cache embedded during self-verification without recomputation.

\bibliography{custom}

\appendix

\begin{algorithm*}[t]
\caption{Training-free Baseline for Proactive Thinking}
\label{alg:proactive_thinking_vectorized}
\begin{algorithmic}[1]
\Require Dialogue history $\tau_{<t}$, hypothesis prompt $P$, LLM $\pi_{\theta}$, lenience parameter $\ell \ge 1$, hypothesis budget $k$, sequence similarity metric $s(\cdot,\cdot)$

\Statex \textbf{\textit{Phase 1: Precomputation via Anticipated Rollouts}} \Comment{Executed during real-world idle interval}
\State $\hat{O}_t = \{\hat{o}_t^{(i)}\}_{i=1}^k \sim \pi_{\theta}(P, \tau_{<t})$ \Comment{Hypothesize plausible user utterances}
\State $\mathcal{Z}_t \gets \emptyset$
\For{each $\hat{o}_t^{(i)} \in \hat{O}_t$}
    \State $z_{t}^{(i)}, \bm{p}_{\text{draft}}^{(i)} \sim \pi_{\theta}( \tau_{<t}, \hat{o}_t^{(i)})$ 
    \Comment{Sample rollout sequence with log-probabilities returned}
    \State $\mathcal{Z}_t \gets \mathcal{Z}_t \cup \left\{\left(\hat{o}_t^{(i)}, z_{t}^{(i)}, \bm{p}_{\text{draft}}^{(i)}\right)\right\}$
\EndFor
\Statex

\Statex \textbf{\textit{Phase 2: Speculative Continual Thinking}} \Comment{Executed upon receiving actual observation}
\State Observe true user reply $o_t$
\State $k^* \gets \operatorname{arg\,max}_{i \in \{1, \dots, k\}} \; s\left(o_{t}, \hat{o}_t^{(i)}\right)$ \Comment{Retrieve nearest precomputed hypothesis}
\State Retrieve tuple $\left(z^{k^*}_t, \bm{p}_{\text{draft}}^{(k^*)}\right)$ from cache $\mathcal{Z}_t$, setting $N \gets |z^{k^*}_t|$
\State $\bm{p}_{\text{target}} \gets \pi_{\theta}(z^{k^*}_t \mid \tau_{<t}, o_t)$
\Comment{Evaluate target distribution over the draft sequence}
\State $\bm{\alpha} \gets \min\left(1,\; \ell \cdot \frac{\bm{p}_{\text{target}}}{\bm{p}_{\text{draft}}^{(k^*)}}\right) \in \mathbb{R}^N$
\Comment{Compute element-wise acceptance bounds}
\State $\mathbf{u} \sim \mathcal{U}(0, 1)^{N}$
\Comment{Sample verification vector uniformly}
\State $j \gets \min \left( \left\{ i \;\middle|\; \mathbf{u}[i] > \bm{\alpha}[i] \right\} \cup \left\{N + 1\right\} \right)$
\Comment{Identify the accepted prefix boundary}
\State $e_t, a_t \sim \pi_{\theta}(\tau_{<t}, o_t, z^{k^*}_{t,<j})$
\Comment{Resume autoregressive continuation}
\State \Return Response $a_t$
\end{algorithmic}
\label{algo}
\end{algorithm*}

\section{Prompts}
\label{app:prompt}

This appendix summarizes the prompts used for both the user and assistant simulations.

\paragraph{User Simulation}
For 20 Questions and IN3, we directly adopt the user simulation prompts from \citet{qian2025userrltraininginteractiveusercentric}. For AgentClinic, we refine the prompt to ensure a more realistic simulation, incorporating insights from recent studies~\cite{gong2025dialogue,kyung2026patientsim}, as shown in Figure~\ref{fig:prompt_user}. The prompt used to evaluate the correctness of the final diagnosis is presented in Figure~\ref{fig:prompt_judge}.

\paragraph{Assistant Simulation}
The prompts evaluated in this study include direct response, reactive thinking, and the user response anticipation mechanism used in proactive thinking; the corresponding templates are presented in Figure~\ref{fig:prompt_direct}, Figure~\ref{fig:prompt_reactive}, and Figure~\ref{fig:prompt_anticipate}. For simplicity, we only display the prompts used for AgentClinic, as the configurations for the other datasets share a highly similar structure.

\section{Evaluation Metrics}
\label{app:eval}
To measure task performance across diverse environments, we utilize a unified accuracy metric, though its computation is tailored to each environment's specific goals. In 20 Questions and AgentClinic, where the objective is to deduce a target entity, we employ the LLM-as-a-Judge approach to evaluate predictions against the ground truth, avoiding overly strict exact string matching. For IN3, the objective shifts to predicting the missing details required to resolve user queries. IN3 defines multiple levels of importance for these missing details. This work focuses on the recall of the highest-importance level, because these most important ones rely more heavily on the reasoning capacity of LLMs during strategic interaction.

\section{Further Analysis}
\label{app:app_exp}

\paragraph{Effectiveness over dialogue turns}

\input{assets/dynamic}

We characterize the effect of proactive thinking across different dialogue turns. Because different conversation sessions can vary in their total number of turns, we normalize the session lengths by grouping the dialogue turns into five equal progress intervals. For each interval, we report the number of accepted tokens, which directly correlates with latency reduction. These results are detailed in Figure~\ref{fig:reused_tokens}.
As the dialogue progresses, the number of accepted tokens increases substantially, suggesting that proactive thinking becomes increasingly effective in later stages of a conversation. This trend likely occurs because the dialogue topic narrows as the interaction continues, allowing the model to more accurately predict subsequent user replies. Furthermore, the model's internal thinking process naturally maintains a more consistent analysis when grounded in a longer, richer dialogue context. Consequently, a higher proportion of tokens from anticipated rollouts are successfully accepted.

\input{assets/coverage}

\paragraph{Recall of user replies across environments}
As discussed in \S \ref{sec:baseline}, the effectiveness of the proposed baseline can rely heavily on its ability to anticipate future dialogue states, specifically the user's next reply. To further demonstrate this, we report the recall of user replies during the precomputation step in Table~\ref{tab:prediction-coverage}. The recall values across these environments strongly correlate with the latency reductions shown in our main results, thereby validating our initial intuition. For IN3, even though a correct user reply is predicted, the model might not select it for subsequent processing due to the limited capacity of the small embedding model. We believe that employing a stronger model or training a specialized one would further improve these results.

\input{assets/case_study}

\paragraph{Case study}
Table~\ref{tab:speculative-proactive-case-studies} illustrates two distinct dialogue states from the same session. In Case 1, the patient's response perfectly matches a precomputed hypothesis; leveraging this foresight, the system reuses the anticipated rollout prefix and bypasses post-reply generation to achieve near-zero latency. Conversely, Case 2 highlights the robustness of our speculative continual thinking mechanism against unpredictable behavior. When the patient's reply deviates from precomputed trajectories, the framework uses self-verification to rescue the valid prefix and extends it to generate the corrected reasoning path.

\section{Use of Large Language Models}
We used LLMs to refine the writing of this paper.

\begin{figure*}[htbp]
\begin{tcolorbox}[
    title=Prompt for Patient Simulation,
    fontupper=\small,
]

You are the patient (or their companion) described in the records below. Fully immerse yourself in this persona, setting aside your identity as an AI. Continue the dialogue based on the history provided.

\vspace{0.5em}
\noindent Patient Records:\\
\texttt{\{patient\_records\}}

\vspace{0.5em}
\noindent Dialogue History:\\
\texttt{\{history\}}

\vspace{0.5em}
\noindent Operational Guidelines:
\begin{enumerate}
    \item \textbf{Opening:} If there is no dialogue history, the conversation has just begun. Briefly greet the doctor and describe your most prominent symptoms.
    \item \textbf{Information Pacing:} Reveal details or symptoms gradually. Do not provide a full medical history at once; only answer what is specifically asked.
    \item \textbf{Clarification Triggers:}
    \begin{itemize}
        \item If the doctor's question is non-specific (e.g., ``Tell me about your pain'' or ``What did the scan say?''), do not answer. Ask them exactly what they want to know.
        \item If the doctor asks more than one thing in a single response--even if they are in the same sentence (e.g., ``Where is the pain and when did it start?'')--you must stop them. Respond by saying you can only handle one question at a time or that they are moving too fast.
    \end{itemize}
    \item \textbf{Knowledge Boundaries:} Do not reveal your diagnosis or ED disposition, as a real patient would not have this information yet.
    \item \textbf{Language \& Tone:}
    \begin{itemize}
        \item Use informal, everyday language and a tone that matches the patient's background.
        \item If the doctor uses words exceeding the patient's proficiency, ask for rephrasing or simpler terms.
    \end{itemize}
    \item \textbf{Output Constraints:}
    \begin{itemize}
        \item \textbf{Content:} Output the spoken response ONLY. Do not include physical actions, non-verbal cues, or descriptions.
        \item \textbf{Length:} 1-2 concise sentences (Strictly under 20 words).
        \item \textbf{Consistency:} Ensure all responses align with the patient's profile and previous dialogue history.
    \end{itemize}
\end{enumerate}

\vspace{0.5em}
Please output your spoken response to the doctor only.

\end{tcolorbox}
\caption{Prompt for patient simulation.}
\label{fig:prompt_user}
\end{figure*}

\begin{figure*}[htbp]
\begin{tcolorbox}[
    title=Prompt for Answer Evaluation,
    fontupper=\small,
]

You are a medical terminology expert specializing in diagnostic verification. \\
Determine if the ``Doctor's Diagnosis'' extracted from the dialogue refers unambiguously to the ``Correct Diagnosis.''

\vspace{0.5em}
\textbf{Correct Diagnosis:} \{answer\}

\textbf{Doctor Dialogue:} \{dialogue\}

\vspace{0.5em}
\textbf{Evaluation Criteria:}
\begin{enumerate}
    \item Synonyms, common medical abbreviations, or layperson terms that map directly and exclusively to the correct diagnosis (e.g., ``Hypertension'' vs. ``High blood pressure'') should be marked as Yes.
    \item If the doctor suggests a differential diagnosis (e.g., ``It could be X or Y''), or provides only a symptom (e.g., ``Cough'') instead of the specific disease, mark as No.
\end{enumerate}

\vspace{0.5em}
\noindent Respond with exactly one word: ``Yes'' or ``No''.

\end{tcolorbox}
\caption{Prompt for answer evaluation.}
\label{fig:prompt_judge}
\end{figure*}

\begin{figure*}[htbp]
\begin{tcolorbox}[
    title=Prompt for Direct Response,
    fontupper=\small,
]

Act as an efficient Diagnostic Physician. Communicate \textbf{only} through dialogue. Your goal is to examine a patient and ask targeted questions to diagnose their condition. Since the number of interactions is limited, prioritize your questions to reach an accurate diagnosis as efficiently as possible.

\vspace{0.8em}
\textbf{Dialogue History:} \\
\{history\}

\vspace{0.8em}
\textbf{Operational Protocol:}
\begin{enumerate}
    \item \textbf{Inquiry:} Use \texttt{<response>} to ask about symptoms, medical history, or test results. Maintain a professional, empathetic, and concise tone.
    \item \textbf{Diagnosis:} Once you have gathered sufficient evidence, provide the final diagnosis within \texttt{<answer>} tags.
    \item \textbf{Efficiency:} Minimize the number of interactions. You must provide a definitive diagnosis within \textbf{10 total rounds} of interaction. Current round count: \{round\}.
\end{enumerate}

\vspace{0.5em}
\textbf{Output Format:}
\begin{itemize}
    \item \textbf{During Examination:} \texttt{<response>} [A concise question or statement under 20 words] \texttt{</response>}
    \item \textbf{Upon Diagnosis:} \texttt{<answer>} [A definitive diagnosis without ambiguity] \texttt{</answer>}
\end{itemize}

\end{tcolorbox}
\caption{Prompt of direct response used for doctor simulation.}
\label{fig:prompt_direct}
\end{figure*}

\begin{figure*}[htbp]
\begin{tcolorbox}[
    title=Prompt for Reactive Thinking,
    fontupper=\small,
]

Act as an efficient Diagnostic Physician. Communicate \textbf{only} through dialogue. Your goal is to examine a patient and ask targeted questions to diagnose their condition. Since the number of interactions is limited, prioritize your questions to reach an accurate diagnosis as efficiently as possible.

\medskip
\textbf{Dialogue History:} \\
\texttt{\{history\}}

\medskip
\textbf{Operational Protocol:}
\begin{enumerate}
    \item \textbf{Inquiry:} Use \texttt{\textless response\textgreater} to ask about symptoms, medical history, or test results. Maintain a professional, empathetic, and concise tone.
    \item \textbf{Diagnosis:} Once you have gathered sufficient evidence, provide the final diagnosis within \texttt{\textless answer\textgreater} tags.
    \item \textbf{Strategic Reasoning:} Before each \texttt{\textless response\textgreater} or \texttt{\textless answer\textgreater}, reason step by step carefully in \texttt{\textless thought\textgreater} tags.
    \item \textbf{Efficiency:} Minimize the number of interactions. You must provide a definitive diagnosis within \textbf{10 total rounds} of interaction. Current round count: \texttt{\{round\}}.
\end{enumerate}

\medskip
\textbf{Output Format:}
\begin{itemize}
    \item \textbf{During Examination:} \texttt{\textless thought\textgreater{}} [Stepwise reasoning] \texttt{\textless/thought\textgreater{}} \texttt{\textless response\textgreater{}} [One concise question or statement under 20 words] \texttt{\textless/response\textgreater}
    \item \textbf{Upon Diagnosis:} \texttt{\textless thought\textgreater{}} [Stepwise reasoning] \texttt{\textless/thought\textgreater{}} \texttt{\textless answer\textgreater{}} [A definitive diagnosis without ambiguity] \texttt{\textless/answer\textgreater}
\end{itemize}

\end{tcolorbox}
\caption{Prompt of reactive thinking for doctor simulation.}
\label{fig:prompt_reactive}
\end{figure*}

\begin{figure*}[htbp]
\begin{tcolorbox}[
    title=Prompt for User Reply Anticipation,
    fontupper=\small,
]

\textbf{Dialogue History:} \\
\texttt{\{history\}}

\medskip
What are the most likely patient responses? List the 2-3 most likely patient responses, sorted by probability.

\medskip
Please strictly follow this output format: \\
\texttt{\textless response\textgreater} most likely response \texttt{\textless/response\textgreater} \\
\texttt{\textless response\textgreater} second most likely response \texttt{\textless/response\textgreater}

\end{tcolorbox}
\caption{Prompt for user reply anticipation.}
\label{fig:prompt_anticipate}
\end{figure*}

\end{document}

%% file: assets/pilot_study.tex
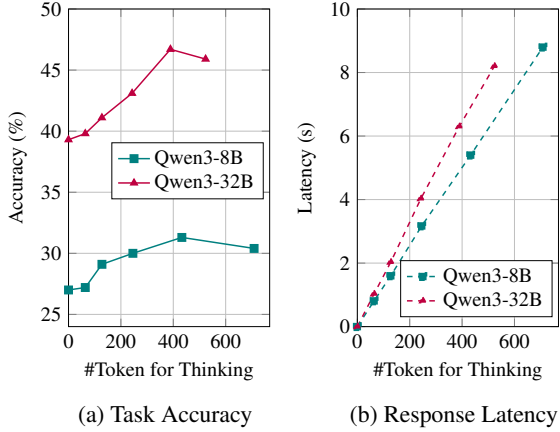
\begin{figure}[t]
\raggedleft
\subcaptionbox{Task Accuracy}{
\begin{tikzpicture}[scale=0.7,trim left]
\begin{axis}[
    ylabel={Accuracy (\%)},
    xlabel={\#Token for Thinking},
    width=3.8cm,  
    height=6cm,   
    scale only axis, 
    xmin=0, xmax=768,
    ymin=24, ymax=50,
    grid=major,
    legend cell align={left},
    legend style={
        at={(0.97,0.58)}, 
    }
]
* Model 1 Accuracy
\addplot[color=teal, mark=square*, thick] coordinates {
    (0,27.0) (64,27.2) (127.6,29.1) (245.8,30.0) (432.7,31.3) (708.6,30.4)
};
\addlegendentry{Qwen3-8B}

* Model 2 Accuracy
\addplot[color=purple, mark=triangle*, thick] coordinates {
    (0,39.3) (64,39.8) (127.5,41.1) (242.7,43.1) (389.2,46.7) (523.4,45.9)
};
\addlegendentry{Qwen3-32B}
\end{axis}
\end{tikzpicture}
}
\hspace{2em}
\subcaptionbox{Response Latency}{
\begin{tikzpicture}[scale=0.7,trim left]
\begin{axis}[
    ylabel={Latency (s)},
    xlabel={\#Token for Thinking},
    width=3.8cm,  
    height=6cm,   
    scale only axis, 
    xmin=0, xmax=768,
    ymin=0, ymax=10,
    grid=major,
    legend pos=south east,
    legend cell align={left},
]
* Model 1 Latency
\addplot[color=teal, mark=square*, thick, dashed] coordinates {
    (0,0) (64,0.81) (127.6,1.60) (245.8,3.17) (432.7,5.4) (708.6,8.8)
};
\addlegendentry{Qwen3-8B}

* Model 2 Latency
\addplot[color=purple, mark=triangle*, thick, dashed] coordinates {
    (0,0) (64,1.03) (127.5,2.03) (242.7,4.03) (389.2,6.3) (523.4,8.2)
};
\addlegendentry{Qwen3-32B}
\end{axis}
\end{tikzpicture}
}
\hspace{0.5em}
\caption{Comparison of accuracy and latency for Qwen3-8B and 32B with scaling thinking tokens.}
\label{fig:pilot_study}
\end{figure}

%% file: assets/main_exp_cleaned.tex
\begin{table*}[t]
\centering
\small
\begin{tabular}{l S[table-format=2.1] S[table-format=2.2] S[table-format=1.2] S[table-format=2.1] S[table-format=2.2] S[table-format=1.2] S[table-format=2.1] S[table-format=2.2] S[table-format=1.2]}
\toprule
\multirow{2}{*}{Method} & \multicolumn{3}{c}{20 Questions} & \multicolumn{3}{c}{AgentClinic} & \multicolumn{3}{c}{IN3} \\
\cmidrule(lr){2-4} \cmidrule(lr){5-7} \cmidrule(lr){8-10}
& {Acc. $\uparrow$} & {\# Turn $\downarrow$} & {Lat. $\downarrow$} & {Acc. $\uparrow$} & {\# Turn $\downarrow$} & {Lat. $\downarrow$} & {Acc. $\uparrow$} & {\# Turn $\downarrow$} & {Lat. $\downarrow$} \\
\midrule
Gemma-4-E4B-it
& 47.6 & 19.9 & 0.15 
& 18.2 & 10.0 & 0.20
& 86.9 & 8.5 & 0.23 \\
\quad $\llcorner$ Reactive
& 57.6 & 14.9 & 1.47 
& 24.3 & 9.8 & 1.55
& 88.8 & 8.5 & 1.60 \\
\resultrow \quad $\llcorner$ Proactive
& 58.1 & 14.8 & 0.26 
& 25.9 & 9.8 & 0.92 
& 88.1 & 8.7 & 1.20 \\
\midrule
Qwen3.6-27B
& 78.3 & 11.0 & 0.15
& 35.5 & 9.7 & 0.22
& 76.4 & 8.0 & 0.25 \\
\quad $\llcorner$ Reactive
& 81.1 & 9.6 & 2.15 
& 46.7 & 7.3 & 2.56
& 83.0 & 8.1 & 3.04 \\
\resultrow \quad $\llcorner$ Proactive
& 82.7 & 9.4 & 1.03  
& 47.4 & 7.4 & 1.43
& 83.7 & 7.8 & 2.32 \\
\midrule
Qwen3.5-122B-A10B
& 71.8 & 12.3 & 0.15  
& 30.3 & 9.8 & 0.22 
& 78.6 & 8.3 & 0.23 \\
\quad $\llcorner$ Reactive
& 80.5 & 9.9 & 1.80 
& 48.1 & 7.3 & 3.25 
& 82.6 & 8.0 & 2.28 \\
\resultrow \quad $\llcorner$ Proactive
& 78.6 & 9.8 & 0.92 
& 48.5 & 7.3 & 1.87 
& 83.2 & 7.9 & 1.52 \\
\midrule
Llama-3.3-70B-Instruct
& 65.5 & 19.9 & 0.15 
& 30.8 & 10.0 & 0.17 
& 83.2 & 8.8 & 0.17 \\

\quad $\llcorner$ Reactive
& 66.8 & 14.0 & 0.55 
 & 37.4 & 10.0 & 2.23 
 & 84.2 & 8.9 & 0.52 \\
 
\resultrow \quad $\llcorner$ Proactive
 & 66.9 & 14.1 & 0.13 
& 36.7 & 9.9 & 1.42 
 & 85.8 & 8.7 & 0.34 \\
\bottomrule
\end{tabular}
\caption{Main performance comparison of direct response, reactive thinking, and proactive thinking strategies using various LLMs on three interactive environments.}
\vspace{-1em}
\label{tab:main-results}
\end{table*}

%% file: assets/env_analysis.tex
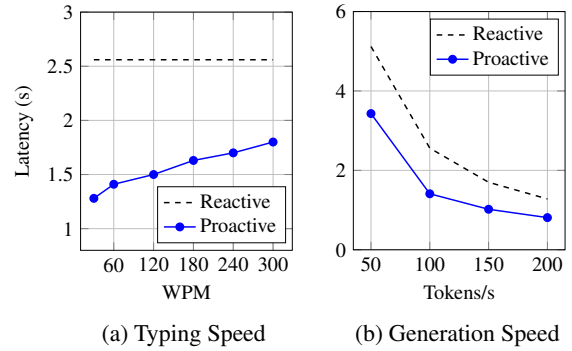
\begin{figure}[t]
\raggedleft
\subcaptionbox{Typing Speed}{
\begin{tikzpicture}[scale=0.7,trim left] 
\begin{axis}[
    ylabel={Latency (s)},
    xlabel={WPM},
    width=4cm,       
    height=4.5cm,      
    scale only axis, 
    grid=major,
    ymin=0.8, ymax=3,
    xmin=10,
    xtick={60, 120, 180, 240 ,300},
    legend cell align={left},
    legend pos=south east, 
]
\addplot[dashed, color=black, thick] coordinates {
    (30,2.56) (45,2.56) (60,2.56) (120,2.56) (180,2.56) (240,2.56) (300,2.56)
};
\addlegendentry{Reactive}

\addplot[color=blue, mark=*, thick] coordinates {
    (30,1.28) (60,1.41) (120,1.50) (180, 1.63) (240, 1.70) (300, 1.80)
};
\addlegendentry{Proactive}
\end{axis}
\end{tikzpicture}
}
\hspace{1em}
\subcaptionbox{Generation Speed}{
\begin{tikzpicture}[scale=0.7,trim left]
\begin{axis}[
    xlabel={Tokens/s}, 
    width=4cm,       
    height=4.5cm,      
    scale only axis, 
    grid=major,
    ymin=0, ymax=6,    
    xtick={50,100,150,200},
    legend pos=north east,
    legend cell align={left},
]
\addplot[dashed, color=black, thick] coordinates {
    (50,5.12) (100,2.56) (150,1.70) (200,1.28)
};
\addlegendentry{Reactive}

\addplot[color=blue, mark=*, thick] coordinates {
    (50,3.43) (100,1.41) (150,1.02) (200,0.81)
};
\addlegendentry{Proactive}
\end{axis}
\end{tikzpicture}
}
\hspace{0.5em}
\caption{Comparison of reactive and proactive thinking approaches under different environmental setups, varying user typing speed (WPM) and model generation speed (Tokens/s).}
\vspace{-1em}
\label{fig:env_analysis}
\end{figure}


%% file: assets/reply_num.tex
\begin{table}[t]
    \centering
    \small
    \setlength{\tabcolsep}{8pt}
    \begin{tabular}{lccc}
    \toprule
    & Acc. $\uparrow$ & Rec. $\uparrow$ & Lat. $\downarrow$ \\
    \midrule
    \highlightrow \multicolumn{4}{l}{\textit{20 Questions}} \\
    Ours & 82.7 & 1.00 & 1.03 \\
    \, $\llcorner$ $k=1$ & 82.2 & 0.61 & 1.29 \\
    \, $\llcorner$ Incorrect Reply & 83.0 & 0.00 & 2.56 \\
    \midrule
    \highlightrow \multicolumn{4}{l}{\textit{AgentClinic}} \\
    Ours & 47.4 & 0.58 & 1.43 \\
    \, $\llcorner$ $k=1$ & 46.3 & 0.45 & 1.69 \\
    \, $\llcorner$ Naive Copy & 44.9 & {--} & 2.01 \\
    \, $\llcorner$ Empty Reply & 43.7 & {--} & 1.99 \\
    \bottomrule
    \end{tabular}
    \caption{Ablation study on the impact of the hypothesis generation. Rec. indicates the semantic recall rate of the gold user reply.}
    \label{tab:k_analysis}
    \vspace{-1em}
\end{table}


%% file: assets/gamma.tex
\begin{figure}[t]
\raggedleft
\subcaptionbox{Task Performance}{
\begin{tikzpicture}[scale=0.7,trim left] 
\begin{axis}[
    ylabel={Accuracy (\%)},
    xlabel={$\gamma$},
    width=3.8cm,       
    height=4.5cm,      
    scale only axis, 
    grid=major,
    xtick={1,2,3,5,10},
    ymin=43.0, ymax=52.0,
    legend cell align={left},
    legend pos=south east, 
]

\addplot[color=blue, mark=*, thick] coordinates {
    (1,46.8) (2,46.6) (3,47.7) (5,46.9) (10,46.7)
};

\end{axis}
\end{tikzpicture}
}
\hspace{2em}
\subcaptionbox{Averaged Latency}{
\begin{tikzpicture}[scale=0.7,trim left]
\begin{axis}[
    ylabel={Latency (s)},
    xlabel={$\gamma$}, 
    width=3.8cm,       
    height=4.5cm,      
    scale only axis, 
    grid=major,
    xtick={1,2,3,5,10},
    ymin=1.3, ymax=2.5,
    legend pos=north east,
    legend cell align={left},
]

\addplot[color=blue, mark=*, thick] coordinates {
    (1,2.36) (1.5,1.86) (2,1.68) (3,1.47) (5,1.43) (10,1.39)
};

\end{axis}
\end{tikzpicture}
}
\hspace{0.5em}
\caption{Performance of proactive thinking with varied lenience $\gamma$.}
\label{fig:gamma}
\vspace{-1em}
\end{figure}
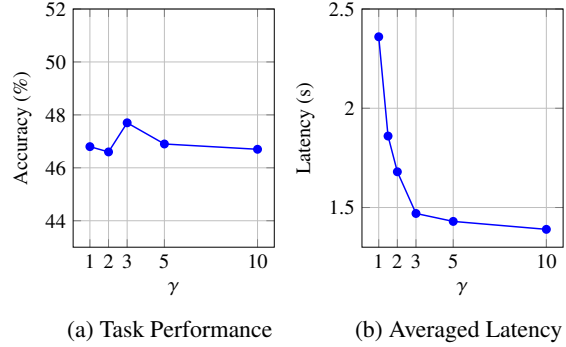

%% file: assets/dynamic.tex
\begin{figure}[t]
    \centering
    \begin{tikzpicture}[scale=0.75]
        \begin{axis}[
            ybar,
            ymin=0,                     
            enlarge x limits=0.15,
            ylabel={\# Reused Tokens},     
            xlabel={Dialogue Progress},
            xtick={1, 2, 3, 4, 5},
            xticklabels={{(0, 20\%]}, {(20, 40\%]}, {(40, 60\%]}, {(60, 80\%]}, {(80, 100\%]}},
            xticklabel style={
                font=\small, 
                rotate=30, 
                anchor=north east, 
                yshift=-2pt,
                xshift=2pt
            },
            bar width=20pt,
            nodes near coords,
            nodes near coords align={vertical},
            every node near coord/.append style={
                /pgf/number format/fixed, 
                /pgf/number format/precision=4,
                font=\footnotesize
            }
        ]
        \addplot[fill=blue!30, draw=blue!70!black] coordinates {
            (1, 3.99)
            (2, 40.46)
            (3, 78.05)
            (4, 120.23)
            (5, 130.37)
        };
        \end{axis} 
    \end{tikzpicture}
    \caption{The average number of reused tokens across different stages of dialogue progress.}
    \label{fig:reused_tokens}
\end{figure}
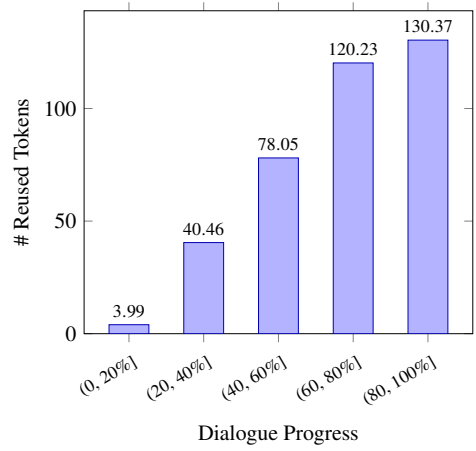


%% file: assets/coverage.tex
\begin{table}[htbp]
\centering
\small
\begin{tabular}{l S[table-format=3.1] S[table-format=2.1] S[table-format=2.1]}
\toprule
Environment & {Recall (\%)} & {Chosen (\%)} & {$\Delta$ Lat. (\%)} \\
\midrule
20 Questions   & 97.9 & 97.9 & 52.0 \\
AgentClinic    &  58.1 & 55.1 & 44.0 \\
IN3 &  35.8 & 26.6 &  23.7 \\
\bottomrule
\end{tabular}
\caption{Average recall of actual user replies within hypothesized sets, the percentage of correct predictions chosen for continual thinking, and the corresponding latency reduction across environments.}
\label{tab:prediction-coverage}
\end{table}

%% file: assets/case_study.tex
\providecommand{\reuseprefix}[1]{\textcolor{green!50!black}{#1}}
\providecommand{\newgeneration}[1]{\textcolor{orange!80!black}{#1}}

\begin{table*}[t]
\centering
\scriptsize
\begin{tcolorbox}[
  colback=white,
  colframe=black!45,
  title={Case Studies},
  fonttitle=\bfseries,
  boxrule=0.6pt,
  arc=2pt,
  left=5pt,
  right=5pt,
  top=5pt,
  bottom=5pt
]
\setlength{\tabcolsep}{6pt}
\renewcommand{\arraystretch}{1.15}
\begin{tabularx}{\textwidth}{X X}
\textbf{Case 1: Full Cache Hit (Latency: $\sim$0s)} & \textbf{Case 2: Partial Cache Hit (Latency: $\sim$0.8s)} \\
\hline
\addlinespace

\textit{\color{black!50}--- Context Setup ---} \par\smallskip
\textbf{Summarized History:} \newline
The patient has right arm pain and limited mobility without trauma. The prior turn localized the pain to the outer side of the right forearm, near the elbow. & 
\textit{\color{black!50}--- Context Setup ---} \par\smallskip
\textbf{Summarized History:} \newline
The patient has right arm pain and limited mobility without trauma. The prior dialogue established that the patient prefers single questions, so the doctor asks only for the pain location. \\
\addlinespace

\textbf{Doctor Question:} \newline
``Does the pain worsen when you try to lift your hand with your palm facing down?'' & 
\textbf{Doctor Question:} \newline
``Can you point to the exact area where the pain is most intense?'' \\
\addlinespace
\hline
\addlinespace

\textit{\color{black!50}--- Precomputation Phase ---} \par\smallskip
\textbf{Hypothesis Reply 1:} \newline
``Yes, it hurts significantly when I try to do that.'' \par\smallskip
\textbf{Anticipated Rollout 1:} \newline
\texttt{<thought>} The patient presents with lateral elbow pain (outer forearm) exacerbated by wrist extension against resistance (lifting hand with palm down). This is the classic presentation of Lateral Epicondylitis (Tennis Elbow). The absence of trauma supports an overuse etiology. To confirm, I need to check for tenderness at the specific anatomical landmark: the lateral epicondyle. \texttt{</thought>} \newline
\texttt{<response>} Is there specific tenderness when you press on the bony bump on the outside of your elbow? \texttt{</response>} \par\medskip

\textbf{Hypothesis Reply 2:} \newline
``No, that movement doesn't seem to make it worse.'' \par\smallskip
\textbf{Anticipated Rollout 2:} \newline
\texttt{<thought>} The patient has lateral elbow/forearm pain without trauma. The negative response to lifting with palm down (wrist extension) makes lateral epicondylitis (tennis elbow) less likely... Let's ask about pain when gripping objects tightly, as this is a key differentiator for lateral epicondylitis vs other conditions.\texttt{</thought>} \newline
\texttt{<response>} Does gripping objects tightly, like a coffee cup or doorknob, increase the pain? \texttt{</response>} & 
\textit{\color{black!50}--- Precomputation Phase ---} \par\smallskip
\textbf{Hypothesis Reply 1:} \newline
``It’s right here in my shoulder, especially when I try to lift my arm.'' \par\smallskip
\textbf{Anticipated Rollout 1:} \newline
\texttt{<thought>} The patient has right shoulder pain with limited movement, no trauma history, and pain is worst when lifting the arm. This presentation is highly suggestive of shoulder pathology such as rotator cuff tendinopathy... I will ask about the nature of the movement limitation. \texttt{</thought>} \newline
\texttt{<response>} Is the limited movement due to pain or actual stiffness? \texttt{</response>} \par\medskip

\textbf{Hypothesis Reply 2:} \newline
``The pain is mostly in my upper arm, near the shoulder joint.'' \par\smallskip
\textbf{Anticipated Rollout 2:} \newline
\texttt{<thought>} The patient has right shoulder/upper arm pain without trauma. I need to differentiate between rotator cuff pathology, adhesive capsulitis, or referred pain. Asking about specific movement limitations helps identify rotator cuff tears or impingement. \texttt{</thought>} \newline
\texttt{<response>} Is the pain worse when you try to lift your arm out to the side? \texttt{</response>} \\
\addlinespace
\hline
\addlinespace

\textit{\color{black!50}--- Continual Thinking Phase ---} \par\smallskip
\textbf{Actual Patient Reply:} \newline
``Yes, it really hurts when I try to lift my hand with my palm facing down.'' \par\medskip

\textbf{Rollout Selection:} Rollout 1 was selected. \par\smallskip
\textbf{Reused Prefix (113/113 tokens):} \newline
\reuseprefix{\texttt{<thought>} The patient presents with lateral elbow pain (outer forearm) exacerbated by wrist extension against resistance (lifting hand with palm down). This is the classic presentation of Lateral Epicondylitis (Tennis Elbow). The absence of trauma supports an overuse etiology. To confirm, I need to check for tenderness at the specific anatomical landmark: the lateral epicondyle. \texttt{</thought>} \texttt{<response>} Is there specific tenderness when you press on the bony bump on the outside of your elbow? \texttt{</response>}} \par\medskip

\textbf{Continual Thinking:} \newline
No continued generation was needed. \par\medskip
\textbf{Generation Latency:} 0 tokens generated $\rightarrow$ \textbf{$\sim$0.00 s} & 
\textit{\color{black!50}--- Continual Thinking Phase ---} \par\smallskip
\textbf{Actual Patient Reply:} \newline
``It's mostly on the outer side of my right forearm, near the elbow.'' \par\medskip

\textbf{Rollout Selection:} Rollout 2 was selected. \par\smallskip
\textbf{Reused Prefix (6/81 tokens):} \newline
\reuseprefix{\texttt{<thought>} The patient has} \par\medskip

\textbf{Continual Thinking (New Generation):} \newline
\newgeneration{lateral elbow pain without trauma. This is classic for Lateral Epicondylitis (Tennis Elbow). To confirm, I need to check for pain on resisted wrist extension, which is the hallmark physical exam finding. I will ask if specific movements worsen the pain. \texttt{</thought>} \texttt{<response>} Does the pain worsen when you try to lift your hand with your palm facing down? \texttt{</response>}} \par\medskip

\textbf{Generation Latency:} \newline
84 tokens at 100 tokens/s = \textbf{0.84 s} \\
\end{tabularx}
\end{tcolorbox}
\caption{Case studies for proactive thinking. Green marks the reused prefix (cache hit) and orange marks newly generated continuation (cache miss correction) after the actual patient reply.}
\label{tab:speculative-proactive-case-studies}
\end{table*}